\documentclass{article}

\usepackage[preprint]{neurips_2019}

\usepackage{graphicx}
\usepackage{subfig}
\usepackage{float}
\usepackage[dvipsnames]{xcolor}

\usepackage[utf8]{inputenc} 
\usepackage[T1]{fontenc}    
\usepackage{hyperref}       
\usepackage{url}            
\usepackage{booktabs}       
\usepackage{amsfonts}       
\usepackage{nicefrac}       
\usepackage{microtype}      
\usepackage{multirow}
\usepackage{amsmath}

\title{ \hspace{0.4in} Recurrent Attentive Neural Process \newline for Sequential Data}
\author{%
  Shenghao Qin$^{1}$\thanks{Authors contributed equally} \hspace{0.3cm} Jiacheng Zhu$^{1*}$  \hspace{0.3cm} Jimmy Qin$^{2}$  \hspace{0.3cm} Wenshuo Wang$^{1}$ \hspace{0.3cm} Ding Zhao$^{1}$ \\
  \text{ } \\
  $^1$Carnegie Mellon University, $^2$ Columbia University \\
  \texttt{shqin16@fudan.edu.cn} \hspace{0.01in} 
  \texttt{jzhu4@andrew.cmu.edu} \\ \texttt{qqin23@gsb.columbia.edu}
   \texttt{ \{wenshuow, dingzhao\}@andrew.cmu.edu} 
}

\begin{document}
\maketitle
\begin{abstract}
Neural processes (NPs) learn stochastic processes and predict the distribution of target output adaptively conditioned on a context set of observed input-output pairs. Furthermore, Attentive Neural Process (ANP) improved the prediction accuracy of NPs by incorporating attention mechanism among contexts and targets.
In a number of real-world applications such as robotics, finance, speech, and biology, it is critical to learn the temporal order and recurrent structure from sequential data. However, the capability of NPs capturing these properties is limited due to its permutation invariance instinct. In this paper, we proposed the Recurrent Attentive Neural Process (RANP), or alternatively, Attentive Neural Process-Recurrent Neural Network(ANP-RNN), in which the ANP is incorporated into a recurrent neural network.
The proposed model encapsulates both the inductive biases of recurrent neural networks and also the strength of NPs for modeling uncertainty. We demonstrate that RANP can effectively model sequential data and outperforms NPs and LSTMs remarkably in a 1D regression toy example as well as autonomous driving applications. 

\end{abstract}

\section{Introduction}

The ordering and recurrent structures of sequential data in many applications, ranging from robotics and finance to speech and biology,  usually carry crucial information. The neural networks based approaches \citep{hochreiter1997long} deal with sequences by costly training on all input-output pairs, thus face obstacles in meta-learning tasks \citep{finn2017meta}. In addition, these models barely capture the propagating uncertainty across time. On the other hand, Gaussian Processes (GPs) \citep{williams2006_GPML}, as a Bayesian nonparametric approach, can model uncertainty flexibly by inferring the distribution over functions; however, it suffers from a high computational burden.
For real-world applications in robotics and autonomous vehicles that involve human interactions, it is essential to efficiently capture not only the representations of underlying temporal dynamics but also the propagating uncertainty from an agent.

Neural Processes (NPs)\citep{garnelo2018conditional,garnelo2018NP} approximate a stochastic process by modelling a distribution over regression functions with prediction complexity linear in the size of observed context set \citep{garnelo2018conditional}. 
NPs estimate an order invariant predictive distribution of \textit{target} output conditioned on  \textit{context} input-output pairs of arbitrary size.
NPs display some of the fundamental properties and capabilities of GPs, but have the weakness of underfitting. Attentive Neural Process (ANP) addressed this drawback by learning the relevant information among contexts and targets via attention mechanisms \citep{kim2019attentive}. As a generalization of NPs, ANP models a stochastic process leveraging on both uncertainty-awareness as Bayesian models and computation efficiency inherited from neural networks. 
Recurrent models with long short-term memory (LSTM) \citep{hochreiter1997long} have been the leading approach for sequential structure modelling at present. 
The LSTM is an efficient gradient-based method for training recurrent networks, which improves the learning process by stabilizing the flow of the back-propagated errors. It defines the state-of-the-art performance on sequential tasks such as speech and text data \citep{graves2013speech_lstm, sutskever2014sequence_text}. \par






In spite of the aforementioned appealing properties, neither NPs nor LSTM models could be solely utilized to properly learn the recurrent structures and propagating uncertainty in temporally ordered sequences. Moreover, given the fact that most real-world engineering problems involve temporal dynamics  that controlled by underlining continuous functions (e.g., Partial Differential Equation (PDE) with time derivatives), the uncertainty also reveals important information such as human factor. It is essential to efficiently capture both the uncertainty and temporal order \citep{al2016gplstm}. While several models are proposed based on state-space models \citep{doerr2018PRSSM}, recurrent neural networks \citep{al2016gplstm}, and Gaussian Processes \citep{wilson2011GP_reg_net}, learning from sequential data remains to be an area of active research.




In this paper, we introduce Recurrent Attentive Neural Process (RANP), or alternatively, Attentive Neural Process Recurrent Neural Network (ANP-RNN) in order to better model continuous sequential data in real-world applications. The proposed ANP-RNN incorporates RNNs into ANP explicitly, thus can capture both the ordering sequential information as well as the propagating uncertainty. 
Benefited by the meta-learning framework of NPs, ANP-RNN learns a stochastic process efficiently from limited observations for multiple tasks. At the same time, it captures the ordering and recurrent features from observed sequences as well.
We demonstrate the effectiveness of ANP-RNN with a 1D function regression task on synthetic data. We further demonstrate that ANP-RNN outperforms NPs, ANP, and LSTMs in trajectory prediction tasks for self-driving applications in terms of accuracy and expressiveness. 

\begin{figure*}[t!]
    \centering
    \includegraphics[width=0.95\linewidth]{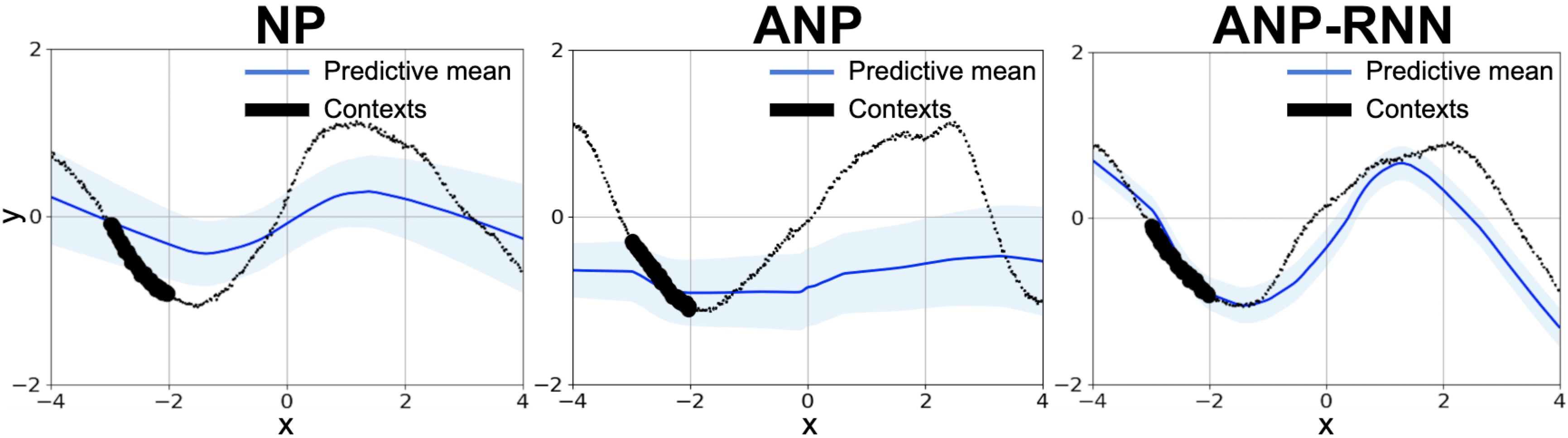}
    \caption{The comparison of predictions given by a fully trained NP, ANP and ANP-RNN in 1D function regression. The contexts (large dots) are used to predict the target outputs ($y$-values of all $x \in [-4,4]$). It is noticeably that the RANP predictions capture both local behaviors from contexts and the global distribution, whereas NP only keeps the global shape of the function and ANP trends to be influenced by contexts dramatically.}
    \label{fig_first_cherry}
\end{figure*}

\section{Background}
\label{background}


\paragraph{Attentive Neural Processes (ANPs)}  The NP is a model for stochastic process realizations that maps an input $X \in \mathbb{R}^{d_x}$ to an output random variable $Y \in \mathbb{R}^{d_y}$. Specifically, NP is defined as a (infinite) family of conditional distributions, in which an arbitrary number of observed \textit{contexts} $(X_C, Y_C):= (x_i, y_i)_{i \in C}$ is used to model an arbitrary number of \textit{targets} $(X_T, Y_T):=(x_i, y_i)_{i \in T}$. Where $C$ depicts a set of $n$ \text{context} points and $T$ describes $m$ unobserved points. The generative process can be written as:
\begin{equation}
P(Y_T|X_T, X_C, Y_C) := \int p(Y_T | X_T, r^*_C, z)q(z | s_C) dz
\end{equation}
where $z$ is a global latent variable describing uncertainty in the predictions of $Y_T$ for a given observation $(X_C,Y_C)$, and is modelled by a factorized Gaussian parameterized by $s_C := s(X_C, Y_C)$, with $s$ being a MLP encoder which represents the \textit{context} $(X_C, Y_C)$ with permutation invariance. 

Meanwhile, deterministic function $r^*_C$ aggregates \textit{contexts} by taking the mean of pair-wise \textit{context} representations in NPs. When in ANP, each target query $X_T$ attends to context by $r^*_C:=r^*(X_C,Y_C,X_T)$ to form query-specific representations. In particular, each context $(x,y)$ pair is passed through an multilayer perceptron (MLP) encoder followed by self-attention layers to form a  pair-wise representations $r_i$, and these are processed in \textit{cross-attention} layers with $X_T$ attended. Under this setting, the global structure of the stochastic process realization is preserved since $z$ induces correlations in the marginal distribution of predictions $Y_T$ in the latent path, where as the fine-grained local structure is captured by the deterministic path. \par
The model parameters are learned via variational approximation by maximizing the following ELBO
\begin{equation}
\log p(Y_T|X_T, X_C,Y_C) \geq \mathbb{E}_{q(z|s_T)}[\log p(Y_T|X_T,r^*_C,z)] - D_{KL}(q(z|s_T) || q(z|s_C))
\end{equation}
using reparameterization trick \citep{kingma2013_AEVB}.
Note that since the conditional prior $p(z|s_T)$ is intractable, the variational posterior $q(z|s_T)$ is used for approximation, as the KL regularization term aligns the summary of \textit{contexts} and \textit{targets} closer to each other. It also reveals the underlying principle of ANPs who learns to infer the \textit{target} stochastic process by assuming that contexts and targets come from the same realization of the data-generating stochastic process, especially when targets contain contexts. 
\par


\paragraph{Recurrent Neural Networks and LSTM} Long Short-Term Memory (LSTM) \citep{hochreiter1997long} is a special form of Recurrent Neural Networks (RNNs) and is one of the most successful ways to exploit sequential information of the data. LSTM places a \textit{memory cell} into each hidden unit and uses a few gate vectors to control the passing of information along the sequence, therefore improve the long-range dependencies and overcome the vanishing gradients problem in RNNs.
The gating mechanism not only improves the flow of errors through time, but also make the network be capable of decide whether to keep, erase, or overwrite memorized information, and therefore increase stability to the network's memory.

\begin{figure*}[t!]
    \centering
    \includegraphics[width=0.95\linewidth]{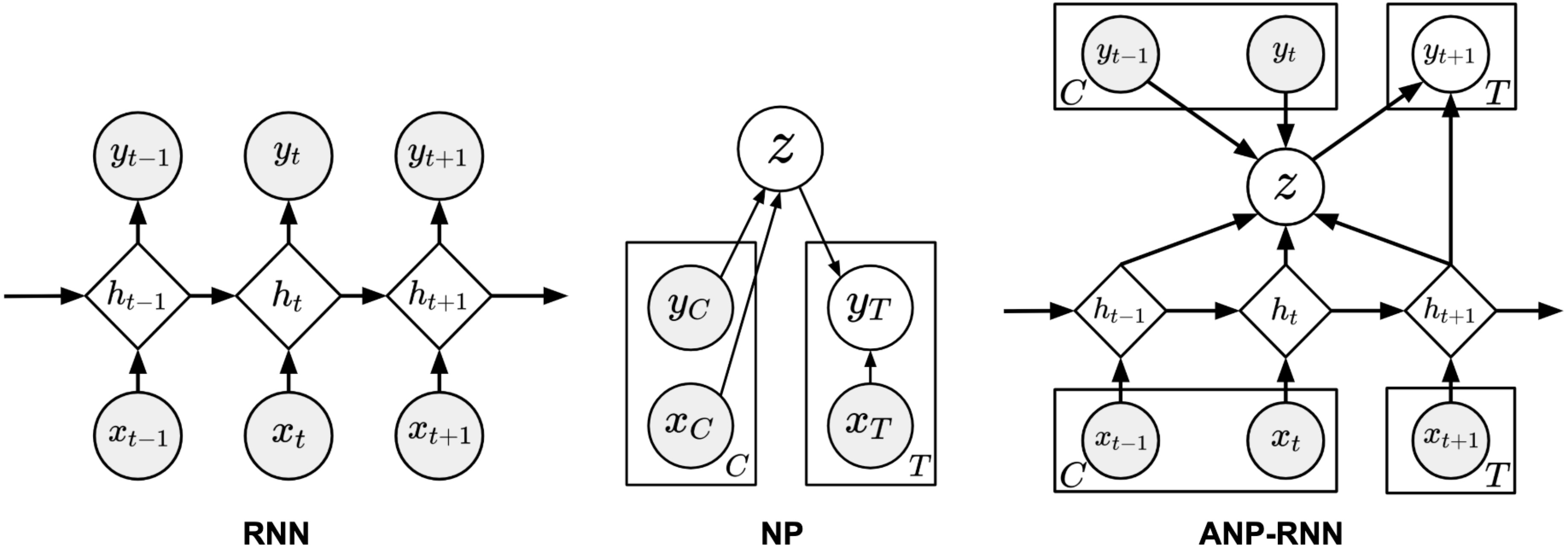}
    \caption{Graphical models of related models and of the ANP-LSTM. Gray shading indicates the observed variables. \textit{C} depicts context variables and \textit{T} for target variables to predict. The diamonds depict deterministic variables.}
    \label{fig_graphical_model}
\end{figure*}

\section{Recurrent Attentive Neural Process}
\label{headings}

In this section, we proposed Attentive Neural Process - Recurrent Neural Network (ANP-RNN) in order to capture ordering and recurrent structures for modeling sequential data. The main idea is to combine the merits of RNNs and ANP.

\begin{figure*}[t!]
    \centering
    \includegraphics[width=\linewidth]{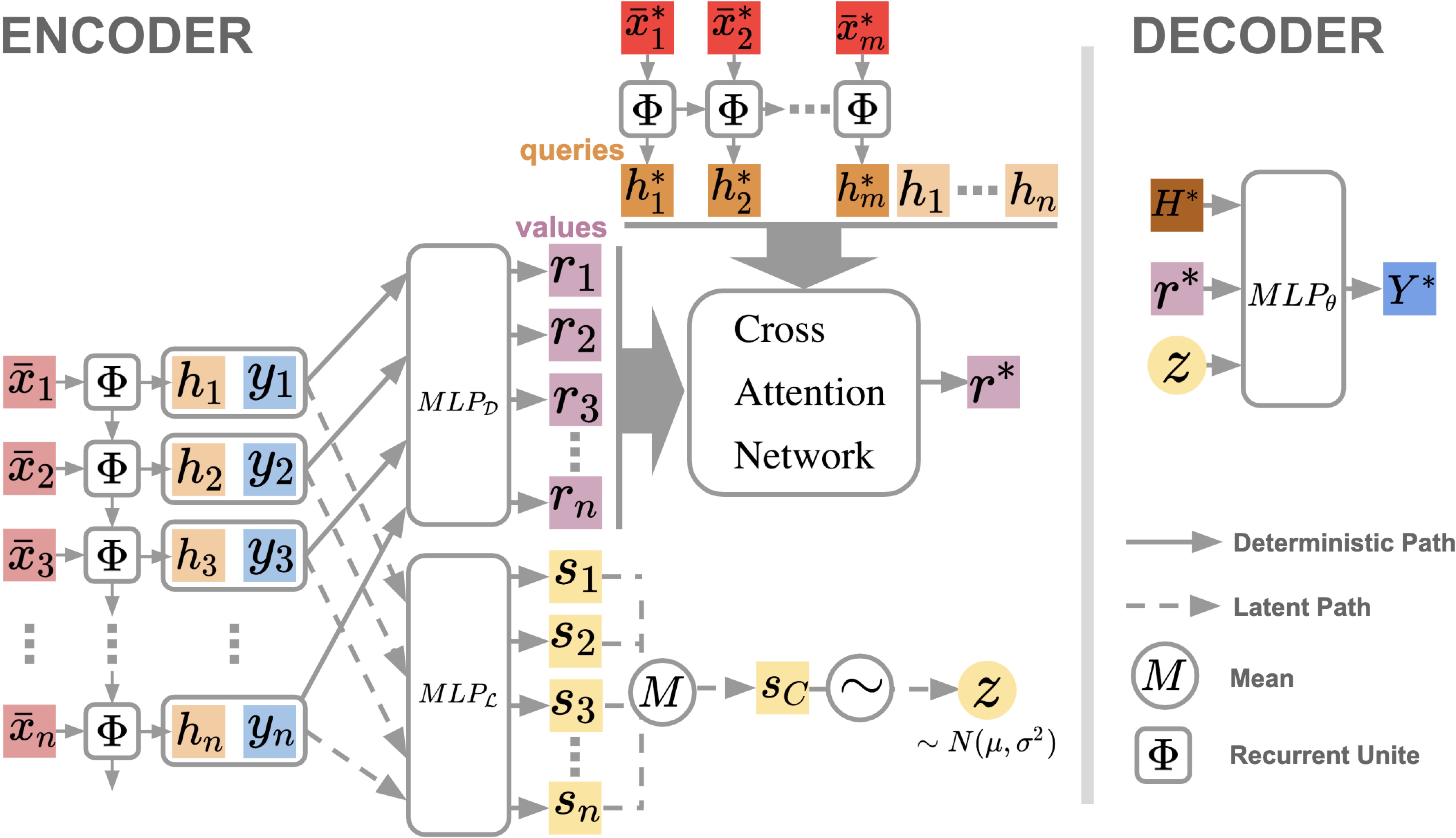}
    \caption{Model architecture of ANP-RNN }
    \label{fig_architecture}
\end{figure*}

\subsection{Model Construction}

We incorporate ANPs with recurrent structure by using an LSTM network to transform the original input space in which the generating stochastic process is modeled, as shown in Figure \ref{fig_architecture}. \par

The sequences of inputs are formally denoted as vectors of measurements $\Bar{x}_1 = [x^1]$, $\Bar{x}_2 = [x^1, x^2],...,\Bar{x}_n=[x^1,x^2,...,x^n]$, $x^t \in \mathcal{X}$, and the length of the sequences $\Bar{x}_n$ would grows. The collection of corresponding real-valued target vectors is $y = \{y_i\}^n_{i=1}$, where $y_i \in \mathbb{R}^d$. It is assumed that only the most recent $L$ steps of a sequence contribute to the prediction of the targets, and the sequences can be written as $\Bar{x}_i = [x^{i-L+1},x^{i-L+2},...,x^i],i=1,...,n$.  Our goal is to model the distribution of functions (realization of stochastic process) $f: \mathcal{X}^L \mapsto \mathbb{R}^d$, which are assumed to be generated from a stochastic process, conditioned on observation data.

Here we first convert an input sequence $\Bar{x}_i$ to a latent representation $h_i \in \mathcal{H}$, and then learn the distribution of realizations that map $h_i$ to $y_i$. Since recurrent models are one of the most promising ways to exploit sequential structure of data. The mapping $f: \mathcal{X}^L \mapsto \mathbb{R}^d$ is represented by recurrent model as
\begin{equation}
\begin{split}
&y_i = \psi(h_i) + \epsilon^t \\
&h^t_i = \phi(h^{t-1}_i, x^{i-L+t}) + \delta^t \text{ ,   }t=1,...,L
\end{split}
\end{equation}
Let the recurrent model $\phi: \mathcal{X}^L \mapsto \mathcal{H}$ be the transformation mapping $\{ \Bar{x}_i \}^n_{i=1}$ to $\{h_i \}^n_{i=1}$, thus the sequential structure in $\{ \Bar{x}_i \}^n_{i=1}$ is integrated into the corresponding latent representations. Subsequently, the ANP is utilized to learn the condition distributions, where the transformed future targets information $(H_T(X), Y_T):=(H_i,Y_i)_{i\in T}=(\Phi(X_i),Y_i)_{i\in T}$ is modelled conditioned on the transformed information of past observed contexts $(H_C(X), Y_C):=(H_i,Y_i)_{i \in C}=(\Phi(X_i),Y_i)_{i \in C}$, in which we certainly have $C \subset T $ from the instinct of time series. Therefore, those conditional distributions are expressed as:
\begin{equation}
p(Y_T | X_T, X_C, Y_C):= \int p(Y_T|H_T, H_C,Y_C,r^*_C,z)q(z|H_C,Y_C) d z
\end{equation}
Here, the mapping from observed inputs to targets vectors is modelled as a stochastic process in a deterministic transferred space. The fact that ANPs treat samples as multiple realizations of a stochastic process enables our proposed model to enlighten the underlying ordering dynamics of sequential data.

\subsection{Learning and Inference}

We train the model using the evidence lower bound (ELBO) of the conditional log likelihood of $Y_T$ as follows

\begin{equation}
\log p(Y_T|X_T, X_C,Y_C) \geq \mathbb{E}_{q(z|s_T)}[\log p(Y_T|X_T,r^*_C,z)] - D_{KL}(q(z|s_T) || q(z|s_C))
\end{equation}

The inference is as follows. First, the input sequences $\{ \bar{x}_i\}^n_{i=1} \in \mathcal{X}^n$ is mapped into representations $\{ h_i\}^n_{i=1} \in \mathcal{H}^n$ using LSTM cells. For notation simplification, we denote $H_C=\Phi (X_C)$ and $H_T=\Phi (X_T)$. Then we can derive the lower bound as \citep{kim2019attentive}
\begin{equation}
\begin{split}
&\log\ p(Y_T|X_T,X_C,Y_C) \geq \\
& \mathbb{E}_{q(z|H_T,Y_T,H_C,Y_C)}[log\ p(Y_T|z,H_T,H_C,Y_C)+ \log\frac{q(z|H_C,Y_C)}{q(z|H_T,Y_T,H_C,Y_C)}]
\end{split}
\end{equation}
where $q(z|H_T,Y_T,H_C,Y_C)$ is represented in latent path as $q(z|s_T)$ and $q(z|H_C,Y_C)$ is represented as $q(z|s_C)$ with MLPs as shown in the graph above.
So we can rewrite the equation as
\begin{equation}
\begin{split}
\log\ p(Y_T|X_T,X_C,Y_C) &\geq \mathbb{E}_{q(z|s_T)}[\log\ p(Y_T|z,H_T,H_C,Y_C)+\log\frac{q(z|s_C)}{q(z|s_T)}]\\
&=\mathbb{E}_{q(z|s_T)}[\log\ p(Y_T|z,H_T,H_C,Y_C)]-D_{KL}(q(z|s_T)||q(z|s_C))
\end{split}
\end{equation}
Finally, we apply the deterministic path representation $r_C^*$ into the equation with $r_C^*:=r(H_C,Y_C)$ containing the information of $H_C$ and $Y_C$ and then get
\begin{equation}
\log\ p(Y_T|X_T,X_C,Y_C) \geq \mathbb{E}_{q(z|s_T)}[\log p(Y_T|H_T,r^*_C,z)] - D_{KL}(q(z|s_T) || q(z|s_C))
\end{equation}

Our model is constructed as simple as possible, meanwhile it is able to represent and quantify \textit{predictive uncertainty} in sequential data. This is achieved by: (1) First, the input sequence is transferred into a latent space by a deterministic LSTM network; (2) Second, the distribution over the functions that map latent sequences to outputs is learned via an ANP. The learning and inference of ANP-RNN/LSTM model is completely achieved by a probabilistic treatment. Thus, the parameters of our proposed model are learned by minimizing negative log-likelihood (NLL), and the prediction are expressed as finding the distribution of targets conditioned on observed contexts.
\par

\section{Related Works}
\label{others}

Recently, there has been a increasing interest in learning and inferring stochastic processes with neural networks. Condition Neural Processes (CNPs) \citep{garnelo2018conditional} model a stochastic process but lack a latent variable for global sampling. Neural Processes (NPs) \citep{garnelo2018NP} models the global uncertainty by introducing an explicit latent path. \citet{kim2019attentive} solved the under-fitting problem and improved the prediction accuracy of NPs by incorporating cross-attention into a latent path. Sequential Neural Processes (SNPs) \citep{singh2019snp} were proposed for modeling non-stationary stochastic process for 4D scene inference with Generative Query Networks (GQN). The Functional Neural Process (FNPs) \citep{louizos2019_FNP} learn the distribution of a function by building a graph of dependencies among local latent variables.    
Generative Query Networks (GQNs) \citep{eslami2018_GQN,kumar2018_CGQN} model the prediction that renders a frame of scene conditioned on a viewpoint, and are regarded as a special case of NPs where $x$ are viewpoints and $y$ are frames. 
\par

Learning from stochastic temporal sequence has been long regarded as a critical problem in the control and dynamical system literature. \textit{State-space models} (SSM) and generative \textit{autoregressive models} are usually applied to describe stochastic temporal processes \citep{van2012SSM_AR}. In the deep stochastic SSM domain, when \citet{chung2015VRNN} added an auto-regressive mechanism to the latent, Deep Kalman Filters (DKF) \citep{krishnan2016DKF} and Deep Variational Bayes Filters (DVBF) \citep{karl2016DVBF} adopted Markovian state transition models for latent states and emission models for observations respectively. Similarly, the above connections can be incorporated into either RNN or Stochastic RNN \citep{biswas2018SRNN}. Recurrent SSM have also been proposed \citep{doerr2018PRSSM} for learning long-term dependencies. Another approach to learn the latent topics in the sequence is developed by combining LSTM with a latent topic model \citep{jo2017_LDA_LSTM} or statistic. Other variants and inference approximations related to modeling stochastic sequences are proposed by \citet{fraccaro2017VAE_LG_SSM} and \citet{krishnan2017NSSM}. In addition, \citet{gemici2017memory_trans} and \citet{fraccaro2018generative_tempoal_model} attach a memory to transition models in order to capture the long-term nonlinear dependencies.

Gaussian Processes (GPs) \citep{williams2006_GPML} are also prevalent in time series modeling. GP-based SSM \citep{eleftheriadis2017_GP_SSM, wang2006_gp_ssm} are introduced in which GPs are adoped as either transition or observation functions. Recurrent GP \citep{mattos2015_RGP} extends the GP-SSM models by applying a recurrent architecture with GP-basd activation functions, however, this model suffers from a sophisticated approximation procedure. Furthermore, GP-LSTM \citep{al2016gplstm} was developed to learn GP kernels with recurrent structure, in which the kernels are learnt via joint optimization for deep networks \citep{wilson2016deepkernellearning, wilson2016stochastic_dkl} and dropout in order to pursue flexibility and scalability \citep{wilson2015_massive_gp}. \par


The most recent related work to our proposed model is GP-LSTM\citep{al2016gplstm}, in which the rare input is embedded by a recurrent neural network for a Gaussian Process regression. However, the sophisticated semi-stochastic block-gradient optimization procedure of GP-LSTM limits its usage. ANP-RNN/LSTM shares a similar philosophy with the GP-LSTM and can learn and predict efficiently due to the neural network structure inherited from NPs. The SNP mentioned above is capable of modeling a non-stationary stochastic process, whereas ANP-RNN/LSTM learns the distribution of a stochastic process by best utilizing the ordering and recurrent information in an observed continues sequence. Moreover, the SNP enhance the latent state $z_t$ by a recurrent structure, while ANP-RNN/LSTM preserves the global latent $z$, as well as the local representation $r^*$, and therefore can better capture the long and short term dependencies in the ordering of a time sequence. 

\section{Experiments}

Since the ANP-RNN learns the stochastic process and is trained on multiple realizations of a stochastic process. At each training iteration, a batch of realizations is drawn from the data generating stochastic process. In order to investigate the ANP-RNN's ability to predict temporal dynamics, with the $x$ axis treated as time index, a sequence of these realizations are selected as contexts and targets. 
The same decoder architecture is used for all experiments, and 8 heads are used for \textit{multihead} attention.



\begin{figure*}[t!]
    \centering
    \includegraphics[width=1.0\linewidth]{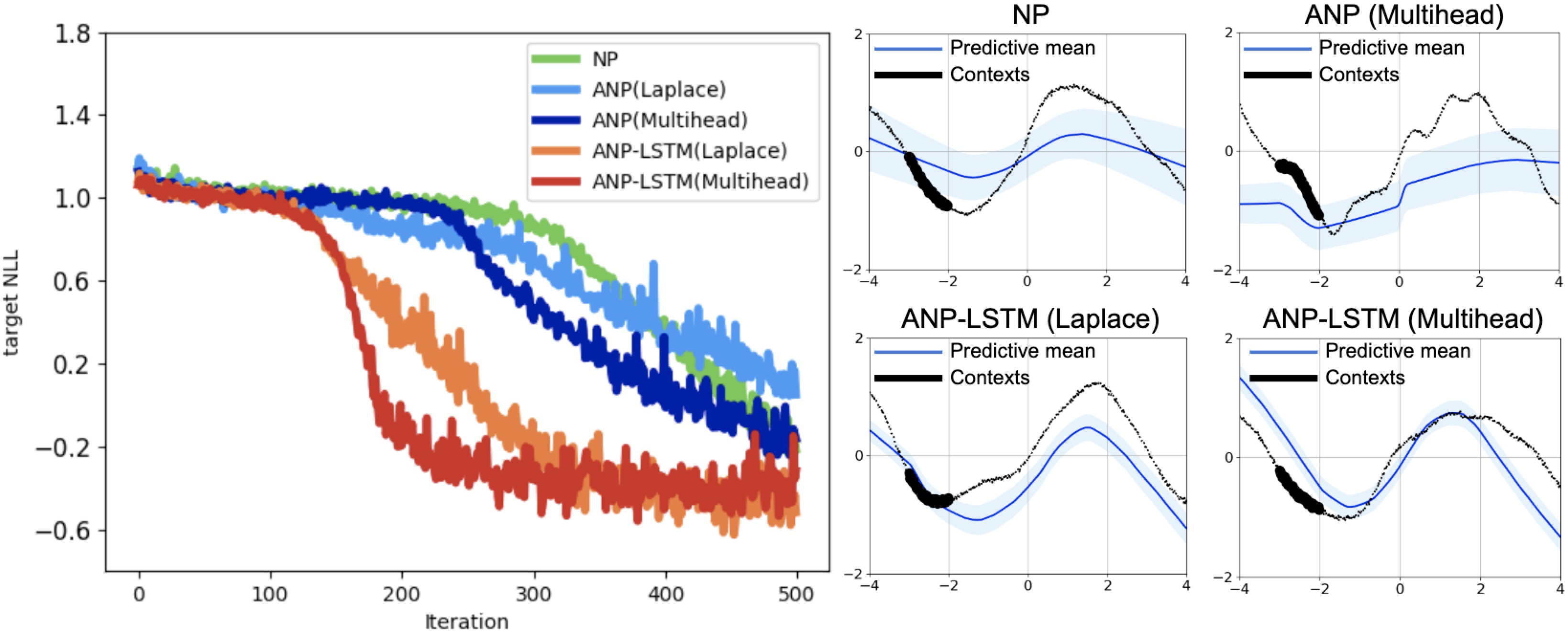}
    \caption{Quantitative comparison between NP, ANP and ANP-LSTM for 1D function regression with random GP kernel hyperparameters. \textbf{Left:} Negative log-likelihood (NLL) for target points given contexts against training iterations for different model settings. \textbf{Right:} Sample predictive mean and variance conditioned on the same context from NP, ANP and ANP-LSTM of different attention mechanisms at iteration $t=400$. The true underlying function is shown in slim black dashed lines while the observed contexts are thick black lines. Best viewed in color.}
    \label{1D_regression}
\end{figure*}

\paragraph{1D Function regression on synthetic stochastic process} The ANP-RNN/LSTM is first tested on data generated from a synthetic stochastic process. In order to illustrate ANP-RNN/LSTM's ability to predict ordered sequences, we construct the stochastic process by adding a Sine function to a Gaussian Process with a squared-exponential kernel and small likelihood noise, rather than just using data from a Gaussian Process. At each iteration, a sequence with fixed length and increment ($X = \{x_i\}^{50}_{i=1},x_{n}=x_0+0.1 \times n$) is chosen from the entire function domain $(\forall n, x_n \in [-4, 4])$ to serve as both contexts and targets. Although NP, ANP, and ANP-RNN/LSTM are all able to make a prediction from permutation invariance contexts\citep{garnelo2018conditional,kim2019attentive}, it is more straightforward to use ordered observations to predict a time series in real-world situations. Here only the cross-attention in the deterministic path is used for ANP and ANP-RNN/LSTM, and we use the same encoder/decoder architecture for NP, ANP, and ANP-RNN/LSTM except for the cross-attention.


\begin{figure}[t!]
    \centering
    \subfloat[Point-wise predictions of the ego-vehicle trajectory by Neural Process (NP) ]{\includegraphics[width=0.95\linewidth]{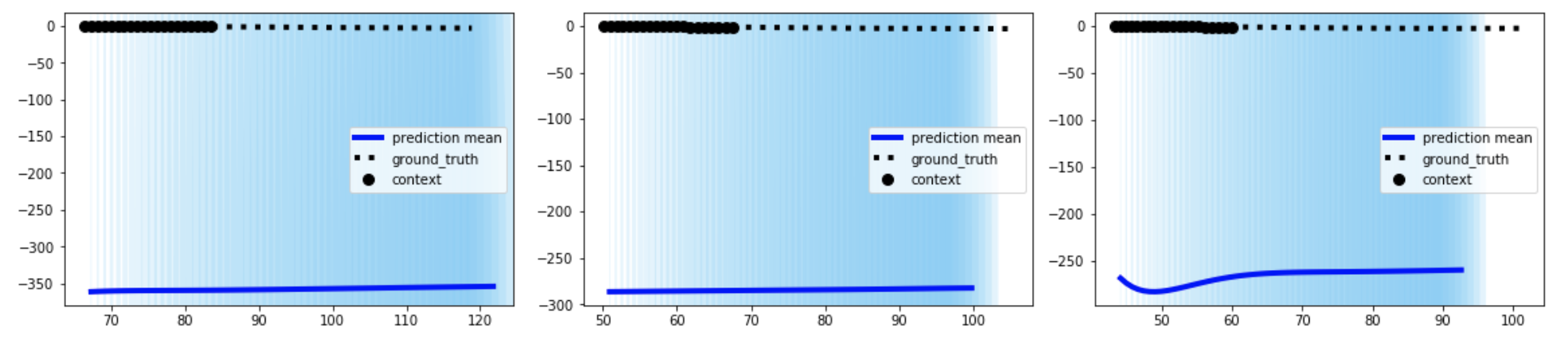}%
    \label{fig2a}}
    \hfil
    \subfloat[Point-wise prediction of the ego-vehicle trajectory by Attentive Neural Process (ANP)]{\includegraphics[width=0.95\linewidth]{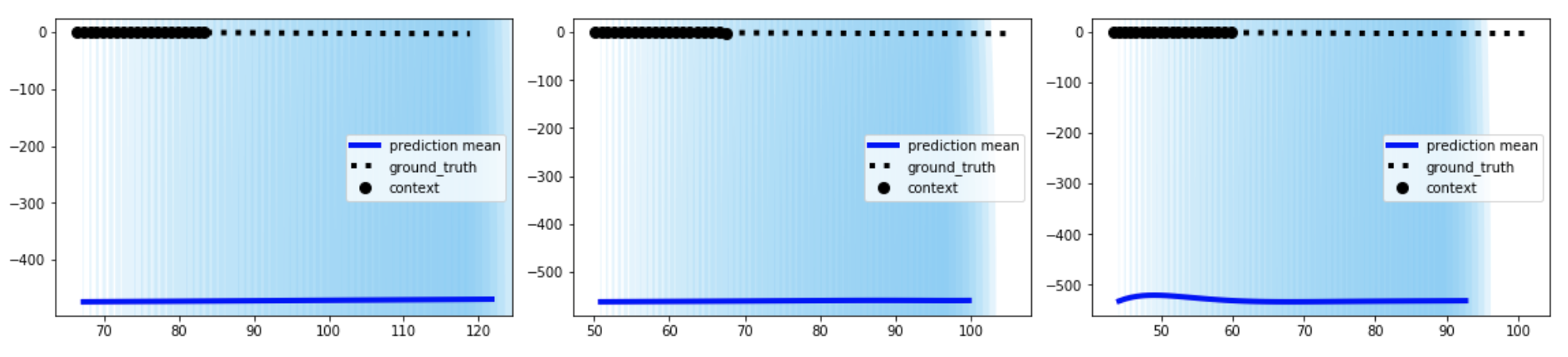}%
    \label{fig2b}}
    \hfil
    \subfloat[Point-wise prediction of the ego-vehicle trajectory by LSTM]{\includegraphics[width=0.95\linewidth]{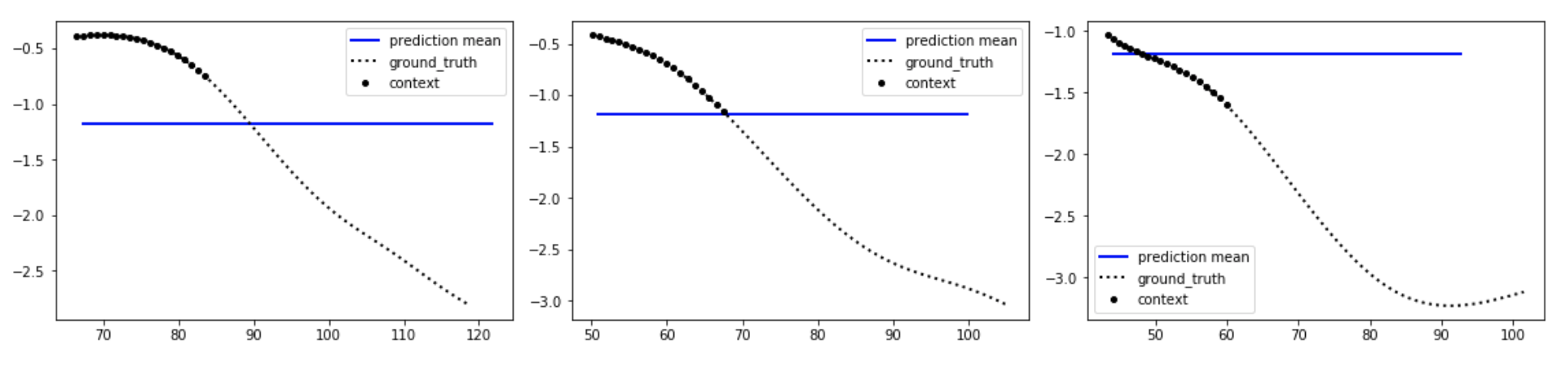}%
    \label{fig2c}}
    \hfil
    \subfloat[Point-wise prediction of the ego-vehicle trajectory by ANP-LSTM ]{\includegraphics[width=0.9\linewidth]{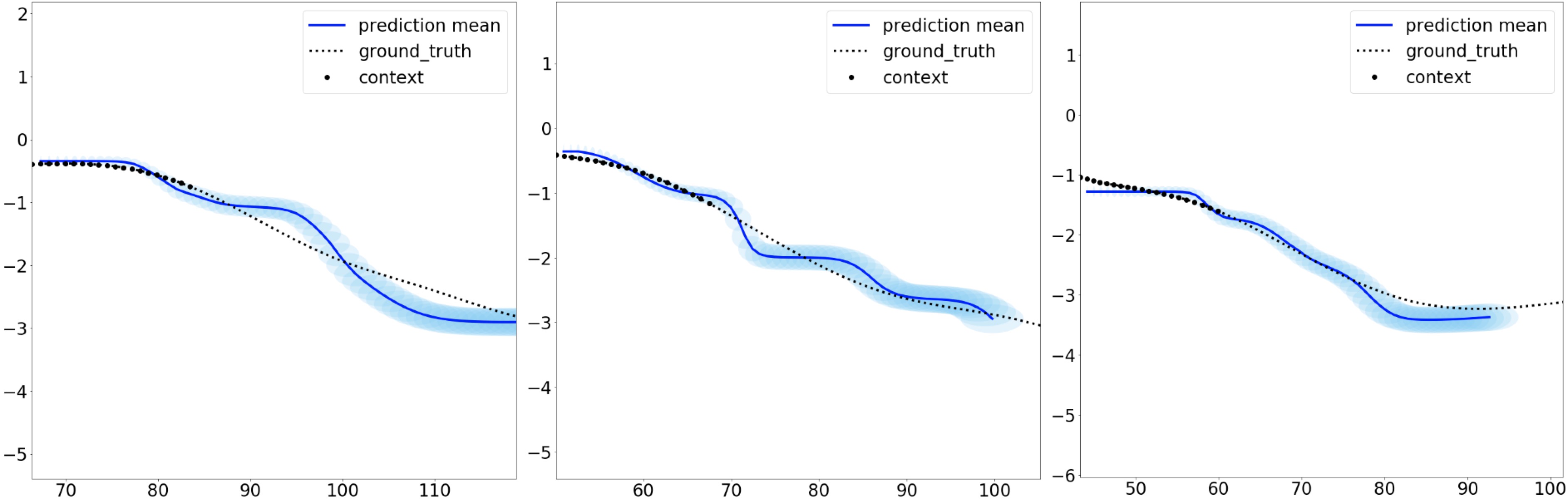}%
    \label{fig2d}}
    \caption{Qualitative comparison of the NP, ANP, LSTM and ANP-LSTM predictions of the lane-changing behavior of autonomous vesicles. Slim dashed lines correspond to the ground truth, blue lines correspond to predictive mean and light blue area depicts the point-wise variance.}
    \label{fig_lane_change}
\end{figure}

Figure \ref{1D_regression} (left) gives negative log-likelihood (NLL) of targets conditioned on giving contents $\frac{1}{|T|}\sum_{i \in T} \mathrm{E}_{q(z|S_C)[\log p(y_i|x_i, r ,z)]}$ for different models trained on a synthetic time series stochastic process. ANP-RNN/LSTM displays a more rapid decrease in NLL compared to (A)NP models, especially for the former with multi-head attention. More qualitative results are provided in Appendix A to demonstrate the performance of different models at different time step. Meanwhile, it can be seen that the multi-head attention mechanism performs better than Laplace attention. In Figure \ref{1D_regression} (right), the learned conditional distributions are shown for a qualitative comparison between different models with different attention mechanisms. The predictive mean of both the NP and ANP with multi-head attention underfits the context. NP attempts to capture the global information of data however results in learning a large likelihood noise, while ANP is sensitive to local dynamics in the data and trends to overfiting certain area. Both ANP-RNN/LSTM with Laplace and multi-head attention mechanism appear to give a good reconstruction of the contexts as well as prediction of contexts. 
Laplace attention is parameter-free (keys and queries are x-coordinates) while the multi-head attention \citep{vaswani2017multihead_attention} makes the query to attend to different keys and thus give smoother query-values \citep{kim2019attentive}. As expected, ANP-RNN/LSTM with multi-head attention gives the best prediction by capture the underlying structure of the synthetic stochastic process.

\paragraph{Autonomous driving traffic scenarios}
In this section the ANP-LSTM is applied to an autonomous driving application along with, Neural Processes (NPs), Attentive Neural Processes (ANPs) and a plain LSTM. We look into the NGSIM dataset and specifically look for lane changing behaviors. In each lane changing scenarios, the target car is going to change its lane in the coming seconds and our goal is to predict its trajectory given the information of its surrounding vehicles. The trajectory of the target vehicle is regarded as a collection of real-value vectors $\mathbf{y}=\{ \mathbf{y_t} \}^n_{t=1} $, $\mathbf{y_t} \in \mathbb{R}^d$, where $n$ is the length of trajectory. The sequence of all surrounding vehicle positions are depicted as vectors of measurements $\boldsymbol{\bar{x}_1} = [\boldsymbol{x^1}] \text{ ,  }  \boldsymbol{\bar{x}_2} = [\boldsymbol{x^1}, \boldsymbol{x^2}], \dots, \boldsymbol{\bar{x}_n} = [\boldsymbol{x^1}, \boldsymbol{x^2}, \dots,\boldsymbol{x^n}]$. Assuming only the most recent $L$ steps of surrounding vehicle positions carries crucial information for predicting target vehicle trajectory, and let  $\boldsymbol{\bar{X}} = \{ {\boldsymbol{\bar{x}_t}} \}^n_{t=1}$ be a collection sequences $\boldsymbol{\bar{x}_i} = [\boldsymbol{x^{i-L+1}}, \boldsymbol{x^{i-L+2}},...,\boldsymbol{x^{i-L+L}}]$ with corresponding length $L$, where $\boldsymbol{x}^i \in \mathcal{X}$. Therefore, we can model and infer the trajectory of interactive vehicle lane changing by model the distribution of a mapping $f:\mathcal{X} \mapsto \mathbb{R}^d$. The difficulty of this tasks lays in the extremely complicated interaction between the target vehicle and surrounding vehicles. Considering that the longitudinal and lateral vehicle dynamics are usually treated separately\citep{rajamani2011_vehicle_dynamics}, our model is trained to predict the two longitudinal and lateral trajectory separately and then jointly predict the positions per frame of the target vehicle.

As attentive neural processes adopts a loss in the form of an approximation of Negative Log-likelihood (NLL), a normal distribution over our target position is given each frame instead of merely a mean prediction. From the Figure \ref{fig_lane_change} we can see that NP, ANP and LSTM cannot capture the hidden structure of the stochastic process and give poor predictions. NP and ANP tend to assign huge standard deviation because of the immense uncertainty of prediction. Thus, the trajectory distribution mean is far from the ground truth. LSTM, as it only predicts a single estimate using mean squared error (MSE), always falls into a local optimum which fails to capture the movement of the vehicle. As for ANP-LSTM, the predicted distribution captures the ground truth closely, indicates that our proposed ANP-LSTM(RNN) is capable of learning and predicting complicated sequential data.
the LSTM part models the time series information into the ANP and then the ANP infers the trajectory using the context information to get quite close prediction curves and velocity fields.


\begin{table}[t]
  \centering
  \begin{tabular}[b]{llllllll}
    \toprule
    \cmidrule(r){1-2}
         & \text{ } 
         & 1s 
         & 2s 
         & 3s
         & 4s
         & MSE
         & NLL\\
    \midrule
    
    \multirow{2}{*}{LSTM} & $\mu$  & 0.286 & -0.330 & -0.588 & -0.776 & 1.4500 & ---\\
                        & $\sigma$ & --- & --- & --- & --- & --- &--- \\[0.2cm]
    \multirow{2}*{NP} & $\mu$ & --- & --- & --- & --- & 92778 & 203882  \\
                        & $\sigma$ & --- & --- & --- & --- & --- & ---\\[0.2cm]
    \multirow{2}*{ANP}    & $\mu$ & --- & --- & --- & --- & 7024.45 & 353693\\
                           & $\sigma$ & --- & --- & --- & --- & --- & --- \\[0.2cm]
    \multirow{2}*{ANP-LSTM} & $\mu$ & \textbf{0.020} & \textbf{0.109} & \textbf{0.130} & \textbf{0.203} & \textbf{0.1673} & \textbf{-0.0229} \\
                       & $\sigma$ & \textbf{0.235} & \textbf{0.276} & \textbf{0.307} & \textbf{0.332} & --- & --- \\
    \bottomrule
  \end{tabular}
  \caption{Absolute Mean Errors and Standard Deviation of the Trajectory Prediction Compared with the Ground-True $\boldsymbol{h}$ in the Lateral Direction. The Unit is in Meters except for the negative log likelihood (NLL). The predictive $\mu$ and $\sigma$ of NP and ANP are omitted since they completely failed model the target vehicle trajectory. }
  \label{sample-table}
\end{table}



\section{Conclusion}

In this paper, Recurrent Attentive Neural Process, or alternatively, ANP-RNN is proposed to model a stochastic process from ordered sequential data. It is demonstrated that this model outperforms NPs, ANPS and LSTM models in terms of accuracy, NLL and qualitative comparison. The ANP-RNN expands the capability of Neural Processes based approaches of learning real-word time series. There exists a large scope of future research for ANP-RNN. Whereas \citet{singh2019snp} added a recurrent in the latent of Neural Process for capture non-stationary dynamics of a stochastic process, their structure could be combined with ANP-RNN in order to better model the complex sequential data from real-world scenarios. Last but the not least, the modest incorporation of NPs and LSTM models makes ANP-RNN open to different variants of recurrent neural networks \citep{oliva2017statisticalRNN} and neural process \citep{singh2019snp,louizos2019_FNP}, and thus bless ANP-RNN with huge potential to solve real-world problems.



\bibliographystyle{plainnat}
\bibliography{main}

\newpage
\section{Appendix A Qualitative Examples for 1D regression \label{appA}}

\begin{figure}[h!]
    \centering
    {\includegraphics[width=0.95\linewidth]{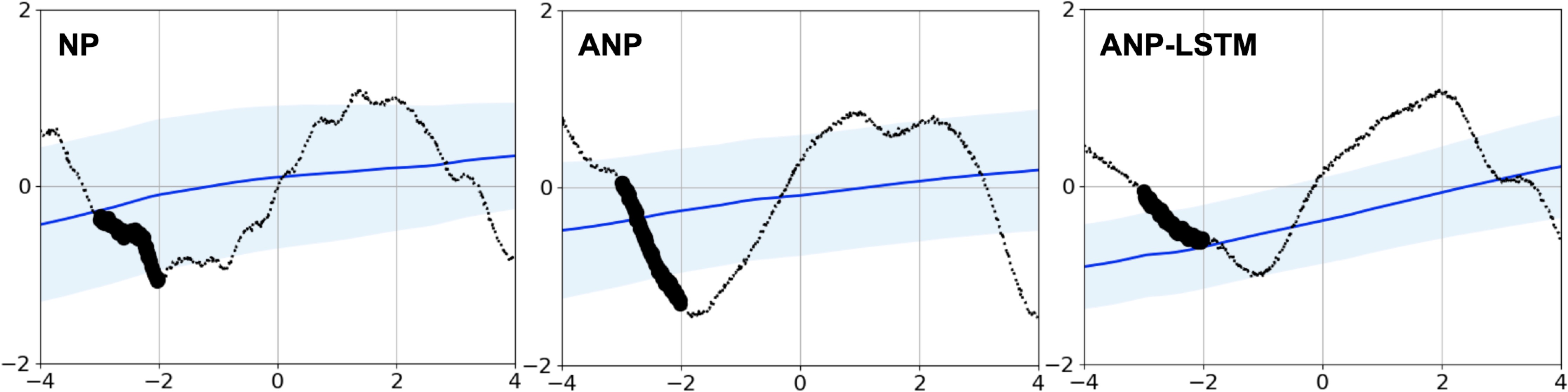}%
    \label{fig_app_a}}
    \hfil
    {\includegraphics[width=0.95\linewidth]{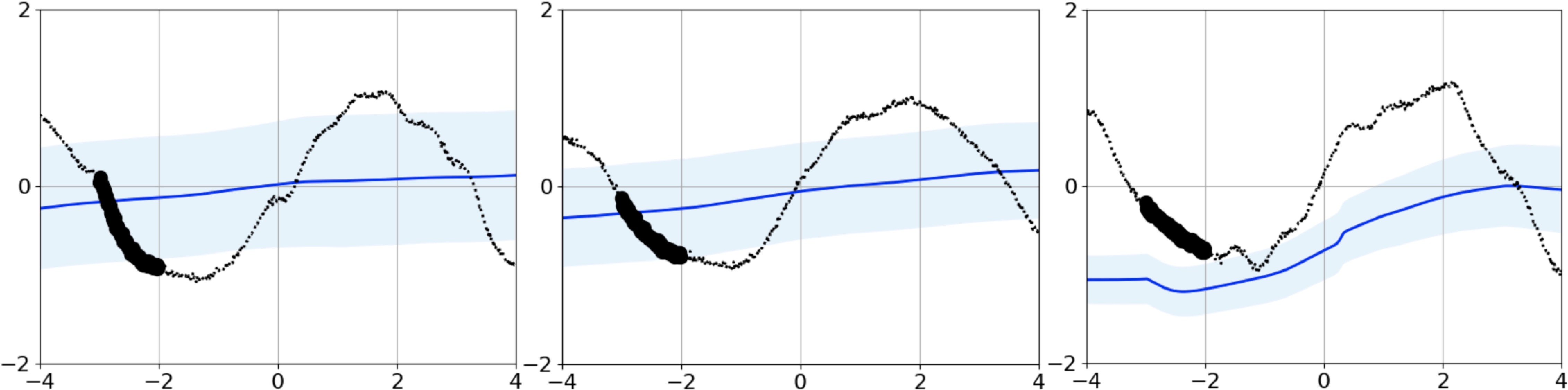}%
    \label{fig_app_b}}
    \hfil
    {\includegraphics[width=0.95\linewidth]{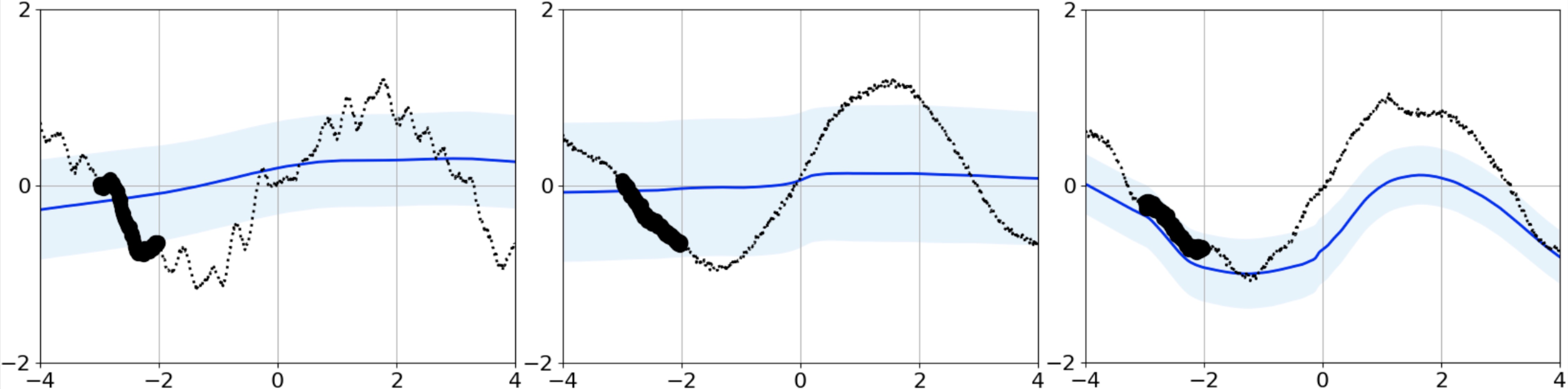}%
    \label{fig_app_c}}
    \hfil
    {\includegraphics[width=0.95\linewidth]{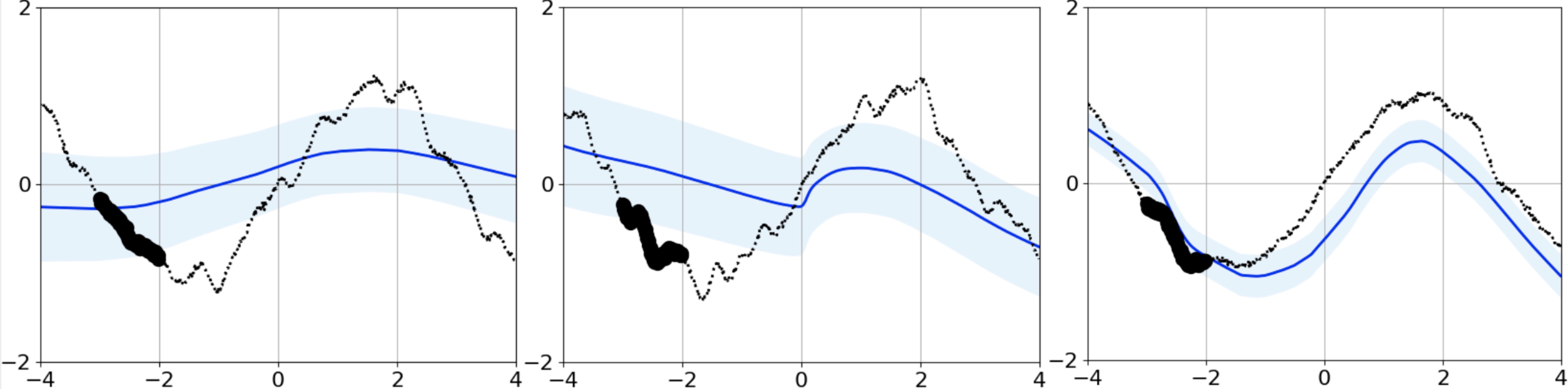}%
    \label{fig_app_d}}
    \hfil
    {\includegraphics[width=0.95\linewidth]{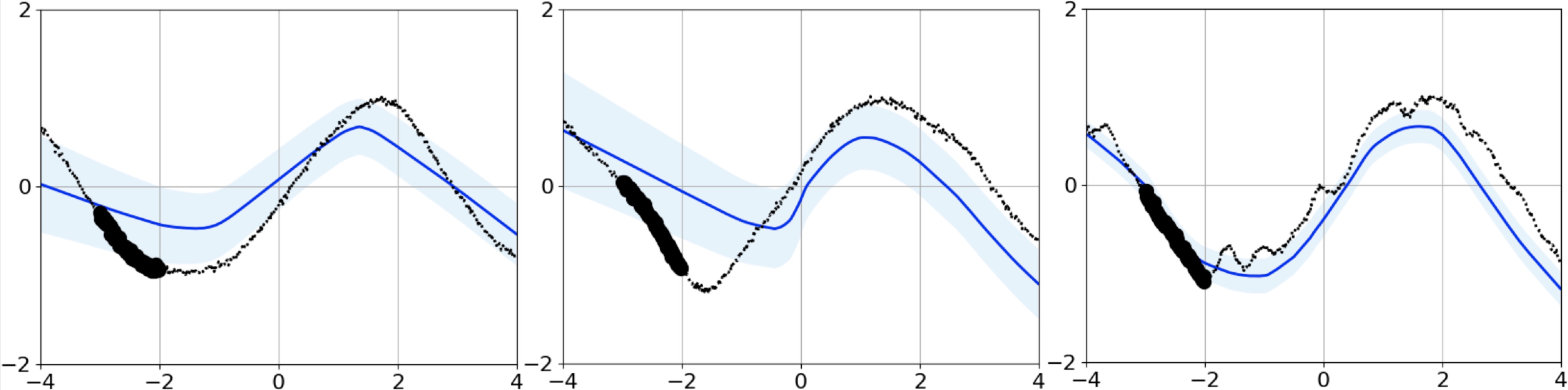}%
    \label{fig_app_e}}
    \caption{Qualitative samples for 1D regression task. Each row corresponds to testing results at time step $t=80,160,240,320,400$ respectively. Each column demonstrates the results generated by NP, ANP, and ANP-LSTM.}
    \label{fig_appendix}
\end{figure}

\end{document}